\documentclass[sigconf,nonacm]{acmart}
\hypersetup{pdfpublisher={Authors}}
\usepackage{colortbl}
\usepackage{arydshln}
\usepackage{placeins}
\AtBeginDocument{%
  }
\usepackage{mdframed}
\begin{document}

\title{Accounting for Irregular State Transitions through Belief Estimation in Online Medical Consultation}

\author{Mingzhe Lu}
\author{Xiangru Chen}
\author{Jian Qin}
\author{Yunpeng Li}
\author{Xihan Gao}
\renewcommand{\shortauthors}{Lu et al.}

\makeatletter
\renewcommand{\@mkauthors}{%
  \global\setbox\mktitle@bx=\vbox{%
    \unvbox\mktitle@bx
    \medskip
    \centering
    {\large\bfseries
      Mingzhe Lu\textsuperscript{1,2}\quad
      Xiangru Chen\textsuperscript{3}\quad
      Jian Qin\textsuperscript{1,2}\quad
      Yunpeng Li\textsuperscript{1,2,*}\quad
      Xihan Gao\textsuperscript{4,*}\par}
    \medskip
    {\normalsize
      \textsuperscript{1}Institute of Information Engineering, Chinese Academy of Sciences, Beijing, China\par
      \textsuperscript{2}School of Cyber Security, University of Chinese Academy of Sciences, Beijing, China\par
      \textsuperscript{3}Culverhouse College of Business, The University of Alabama, Tuscaloosa, AL, USA\par
      \textsuperscript{4}Business Administration, Chang'an University, Xi'an, Shaanxi, China\par}
    \smallskip
    {\small\textsuperscript{*}Corresponding authors: \href{mailto:liyunpeng@iie.ac.cn}{\texttt{liyunpeng@iie.ac.cn}}; \href{mailto:xihangao@chd.edu.cn}{\texttt{xihangao@chd.edu.cn}}.\par}
    \bigskip}}
\makeatother

\begin{abstract}
While online medical consultation broadens access, unprompted patient disclosures and topic shifts frequently disrupt clinician inquiry, leaving clinical states progressively revealed.
We characterize these dynamics using both existing evaluation metrics and task-specific interaction metrics designed in this work.
Through an empirical study across 16 medical dialogue datasets, we show that authentic online interactions exhibit substantially lower contextual continuity and higher unpredictability than offline and simulated benchmarks.
We propose PRISM (Policy Reasoning with Inferred States for Medicine), a framework inspired by partially observable Markov decision processes that decouples latent state estimation from action selection to address drifting dialogue trajectories. Specifically, PRISM factorizes clinical triage into belief estimation, policy optimization, and response realization. By maintaining an evolving posterior belief over consultation states and conditioning action selection on the maximum a posteriori state, PRISM advances clinical goals while accommodating patient-initiated shifts.
Extensive evaluations show that PRISM substantially outperforms frontier LLMs and clinical agents, achieving a 22.40 percentage point gain in clinical decision accuracy alongside decisive improvements in response generation, surpassing the strongest baseline by 5.51 points in BERTScore\_F1.
Furthermore, ablation studies show that providing oracle consultation states consistently boosts response quality across all evaluated baselines by guiding their action reasoning, yielding average gains of 5.58 points in BERTScore\_F1.
\end{abstract}
\maketitle

\section{Introduction}
\textit{Relevance to Web.} Online medical consultation has substantially broadened patient access to professional healthcare~\cite{hwang2022teleconsultations}. However, managing the surging volume of patient inquiries frequently encroaches on physicians' off-hours, imposing a substantial burden that spurs the growing need for intelligent clinical triage~\cite{mandal2024work}.
To effectively deliver such assistance, one must first account for how the online medium reshapes clinical communication. Traditional encounters follow a physician-led inquiry, but online platforms empower patients to drive the conversation at their own pace. Patients can freely volunteer additional symptoms, bring up secondary anxieties, or pivot to a new topic while preceding clinical questions remain unanswered.
This interactional reality prompts us to re-examine whether current intelligent triage systems adequately account for such online uncertainties in user modeling, much less incorporate this understanding into personalized clinical decision-making.

Recent research has increasingly turned to multi-turn interactions for information elicitation and iterative clinical reasoning.
ClinDiag-GPT, for instance, constrains patient agents to provide medical details strictly upon direct physician inquiry~\cite{chen2026grounding}. Even when systems like AMIE permit patients to volunteer supplementary concerns, the interactions remain firmly anchored to predefined clinical profiles and diagnostic goals~\cite{tu2025towards}.
In authentic online consultations, however, patient behaviors are far more complex. 
As ThReadMed-QA demonstrates~\cite{munnangi2026threadmed}, medical evaluations predominantly rely on idealized conditions, nudging most methods to merely cater to exam-style benchmarks. 
Worse still, MedDialBench~\cite{luo2026meddialbench} shows that even advanced models degrade sharply when faced with uncooperative patient utterances.

Although authentic corpora exist, prevailing studies largely focus on specific downstream applications, such as entity recognition~\cite{ctdd}, empathetic counseling~\cite{psychdialog}, or disease-specific archiving~\cite{diabetescare}, thereby exempting current systems from grappling with genuine patient interaction dynamics.
Even recent benchmarks like ThReadMed-QA~\cite{munnangi2026threadmed} empirically uncover these distinct patient behaviors and model vulnerabilities, but stop short of formulating dialogue strategies to resolve them.
We address these limitations by modeling clinical triage under partial observability, where belief estimation is decoupled from sequential action selection.
Before operationalizing this formulation, however, we must systematically analyze the very nature of this partial observability, unpacking whether it is indeed pervasive and what interactional behaviors constitute it.

To this end, we assemble 15 public medical dialogue datasets released between 2023 and 2026, analyzing them alongside our corpus of real online consultations. Together, these datasets cover all four combinations of interaction type (real vs.\ simulated) and clinical setting (online vs.\ offline).
We first evaluate patient messages at the utterance level, revealing that they are inherently fragmented as asynchronous online communication frequently involves delayed and interleaved responses.
Scaling to the dialogue level, we examine both patient initiative behaviors and how physicians accommodate these unprompted contributions.
Finally, at the corpus level, we show that these interaction patterns exhibit distributional cohesion across real-world datasets.
These empirical analyses demonstrate that the online medical consultation state evolves through irregular state transitions, motivating us to explicitly infer evolving belief distributions before selecting triage actions.

We introduce PRISM (Policy Reasoning with Inferred States for Medicine), an interaction framework that decouples belief estimation from state-conditioned policy optimization under partial observability.
Specifically, it tracks an evolving posterior belief distribution over plausible consultation states, utilizing the maximum a posteriori (MAP) state to sequentially condition action selection and response generation.
This explicit factorization enables the model to appropriately handle patient-initiated topic shifts, thereby keeping the consultation on track.
By grounding each response in an updated interpretation of the consultation, PRISM aims to maintain diagnostic momentum amid patient-driven shifts, thereby enhancing the overall efficiency and reliability of intelligent clinical triage.

Our contributions are as follow.
\begin{itemize}
  \item First, we empirically analyze interactional dynamics across 15 public medical dialogue datasets alongside our real-world online corpus. We show that frequent unprompted disclosures and topic shifts amid pending inquiries distinguish authentic online consultations from offline and simulated encounters.
  \item Second, we propose PRISM, a framework that draws inspiration from partially observable decision processes to decouple latent belief estimation from state-conditioned policy optimization. By conditioning action selection and response generation on the MAP state, it accommodates patient topic shifts while pursuing clinical goals.
  \item Third, we extensively evaluate PRISM against frontier LLMs, advanced agents, and specialized medical systems. Specifically, it achieves a 22.40 percentage point gain in structured decision accuracy over the strongest baseline. Crucially, systematic ablations demonstrate that introducing these localized consultation states consistently boosts response performance across all evaluated baselines, achieving an average gain of 5.58 points in BERTScore F1.
\end{itemize}

\section{Research Questions}
We study online medical consultation, focusing primarily on intelligent clinical triage. Clinical triage serves to assess initial patient conditions and guide care pathways, where the primary objective is to systematically elicit structured clinical information and history. In this setting, LLMs conduct multi-turn patient interactions to collect necessary clinical facts, aiming to alleviate physician workload without incurring heavy supervisory overhead.

However, real-world triage dialogues are inherently dynamic and patient-led. Specifically, patients frequently volunteer unsolicited details, leave prior inquiries unanswered, or introduce new concerns that shift the clinical trajectory. More critically, patients often seize the conversational initiative by raising diverse inquiries regarding disease prognosis, medication adjustments, clinic operating hours, and medical expenses. This behavior pushes the already non-linear triage process into an even more intricate interaction scenario heavily interweaved with ad-hoc question answering.

Specifically, at decision step $t$, let $u_t$ denote the latest patient input, which may comprise multiple consecutive messages, and let $y_t$ denote the assistant's patient-facing response. The observable interaction history is $h_t = (y_0, u_1, y_1, \ldots,\allowbreak u_{t-1}, y_{t-1}, u_t)$, where $y_0$ is an optional initial physician utterance. The underlying clinical state $s_t \in \mathcal{S}$, which represents the current consultation focus and its associated information requirements, is only partially observable through $h_t$, with uncertainty represented by a belief distribution $b_t(s) = P(s_t = s \mid h_t)$.
Based on the observed history, the assistant selects an action $a_t \in \mathcal{A}$ to advance the consultation, clarify information, or triage to a physician, aiming to maximize task progress while minimizing physician workload.

Understanding and improving this decision process requires examining how interaction dynamics unfold in practice and affect model behavior. We organize our study around three research questions (RQs).
\begin{itemize}
    \item \textbf{RQ1.} What distinctive interaction patterns characterize authentic online medical consultations, and how pervasively do they manifest?
    \item \textbf{RQ2.} How effectively does state guidance support clinical action selection in unstructured online consultations?
    \item \textbf{RQ3.} How effectively and universally does conditioning on these consultation states enhance response quality across diverse intelligent triage systems?
\end{itemize}

\section{Empirical Study}
To address \textbf{RQ1}, we conduct an empirical study across 15 public medical dialogue datasets and our real-world online consultation corpus, covering diverse online, offline, simulated, and authentic clinical settings. Key characteristics and comparative statistics across all datasets are summarized in Table~\ref{tab:rq1-results}.

\subsection{Setup}
We comprehensively surveyed medical dialogue datasets released between 2023 and 2026. Excluding those requiring credentialed access or restricted licenses, we finalized 15 publicly accessible corpora alongside our authentic online consultation corpus, covering diverse online, offline, simulated, and authentic clinical settings.

These 15 corpora span all four combinations of communication channel and interaction mode:
(1)~\emph{Simulated offline} datasets reflect traditional in-person clinical workflows, focusing on clinical note generation in ACI-Bench~\cite{acibench} and NoteChat~\cite{notechat}, history-taking in MediTOD~\cite{meditod}, and psychiatric or complex consultation scenarios in MDD-5k~\cite{mdd5k}, MedSynth~\cite{medsynth}, and MedDialBench~\cite{luo2026meddialbench}.
(2)~\emph{Real offline} corpora are rarely publicly available owing to privacy and consent constraints. Although some datasets have been publicly released, given licensing restrictions, COMMUNICARE~\cite{communicare} was the only eligible corpus that we could find.
(3)~\emph{Simulated online} benchmarks extend preset cases and patient profiles into interactive exchanges, as demonstrated in MidMed~\cite{midmed}, DISC-Med-SFT~\cite{discmed}, MedPI~\cite{medpi}, and AMIE~\cite{lievin2026conversational}.
(4)~\emph{Real online} resources encompass specialized telemedicine interactions, including psychological counseling in PsychDialog~\cite{psychdialog}, diabetes management in DiabetesCare~\cite{diabetescare}, and general online consultations in CTDD~\cite{ctdd} and ThReadMed-QA~\cite{munnangi2026threadmed}, which we examine alongside our real-world online corpus.

\begin{table*}[t]
    \centering
    \caption{Interaction dynamics of the evaluated samples (rates in $[0,1]$, BPB in bits/byte). $^{*}$ denotes Chinese corpora. Arrows indicate support directions. Shaded rows report group means.}
    \label{tab:rq1-results}
    \begingroup
    \setlength{\tabcolsep}{4.8pt}
    \begin{tabular}{lllrrrrrrrrrr}
        \toprule
        & \multicolumn{2}{c}{Attributes} & \multicolumn{2}{c}{Samples used} & \multicolumn{2}{c}{Utterance} & \multicolumn{3}{c}{Discourse (Patient)} & \multicolumn{2}{c}{Discourse (Physician)} & Corpus \\
        \cmidrule(lr){2-3}\cmidrule(lr){4-5}\cmidrule(lr){6-7}\cmidrule(lr){8-10}\cmidrule(lr){11-12}\cmidrule(lr){13-13}
        Dataset & Type & Mode & \multicolumn{1}{c}{Dial.} & \multicolumn{1}{c}{Utt.} & \multicolumn{1}{c}{ECoh$\,\downarrow$} & \multicolumn{1}{c}{BPB$\,\uparrow$} & \multicolumn{1}{c}{Discl.$\,\uparrow$} & \multicolumn{1}{c}{Redir.$\,\uparrow$} & \multicolumn{1}{c}{Any$\,\uparrow$} & \multicolumn{1}{c}{Cond.$\,\uparrow$} & \multicolumn{1}{c}{Colla.$\,\uparrow$} & \multicolumn{1}{c}{MAUVE$\,\uparrow$} \\
        \midrule
        MedSynth & Sim. & Off. & 5.00k & 242.38k & 0.924 & / & 0.044 & 0.007 & 0.050 & 0.375 & 0.019 & / \\
        ACI-Bench & Sim. & Off. & 0.21k & 11.21k & 0.809 & / & 0.079 & 0.013 & 0.086 & 0.387 & 0.034 & / \\
        MediTOD & Sim. & Off. & 0.21k & 20.31k & 0.905 & / & 0.079 & 0.003 & 0.081 & 0.457 & 0.035 & / \\
        NoteChat & Sim. & Off. & 5.00k & 142.90k & 0.903 & / & 0.088 & 0.005 & 0.091 & 0.432 & 0.040 & / \\
        MDD-5k$^{*}$ & Sim. & Off. & 4.63k & 214.36k & 0.932 & 0.761 & 0.168 & 0.004 & 0.168 & 0.256 & 0.043 & 0.005 \\
        MedDialBench & Sim. & Off. & 5.00k & 129.93k & 0.718 & / & \underline{\textit{0.352}} & 0.120 & 0.407 & 0.551 & 0.235 & / \\
        \rowcolor{gray!12}
        \multicolumn{3}{l}{\textit{Mean}} & / & / & 0.865 & 0.761 & 0.135 & 0.025 & 0.147 & 0.410 & 0.068 & 0.005 \\
        \midrule
        COMMUNI.CARE & Real & Off. & 0.03k & 4.68k & 0.694 & / & 0.105 & 0.018 & 0.116 & 0.592 & 0.069 & / \\
        \rowcolor{gray!12}
        \multicolumn{3}{l}{\textit{Mean}} & / & / & 0.694 & / & 0.105 & 0.018 & 0.116 & 0.592 & 0.069 & / \\
        \midrule
        AMIE & Sim. & On. & 0.36k & 6.46k & 0.669 & / & 0.141 & 0.092 & 0.212 & 0.780 & 0.134 & / \\
        MidMed$^{*}$ & Sim. & On. & 5.00k & 56.61k & 0.860 & 1.272 & 0.076 & 0.027 & 0.090 & 0.567 & 0.050 & 0.054 \\
        MedPI & Sim. & On. & 5.00k & 74.97k & 0.973 & / & 0.191 & 0.052 & 0.214 & \textbf{0.909} & 0.179 & / \\
        DISC-Med-SFT$^{*}$ & Sim. & On. & 5.00k & 36.03k & 0.608 & 1.035 & 0.189 & 0.132 & 0.249 & \underline{\textit{0.898}} & 0.222 & \underline{\textit{0.494}} \\
        \rowcolor{gray!12}
        \multicolumn{3}{l}{\textit{Mean}} & / & / & 0.778 & 1.154 & 0.149 & 0.076 & 0.191 & 0.789 & 0.146 & 0.274 \\
        \midrule
        CTDD$^{*}$ & Real & On. & 0.10k & 1.45k & \textbf{0.415} & 1.385 & 0.276 & 0.172 & 0.342 & 0.621 & 0.185 & 0.322 \\
        PsychDialog$^{*}$ & Real & On. & 3.05k & 21.08k & 0.581 & 1.373 & 0.304 & 0.175 & 0.373 & 0.513 & 0.189 & 0.395 \\
        ThReadMed-QA & Real & On. & 0.10k & 0.72k & 0.679 & / & \textbf{0.436} & 0.082 & \textbf{0.449} & 0.544 & 0.251 & / \\
        DiabetesCare$^{*}$ & Real & On. & 3.48k & 32.73k & 0.600 & \textbf{1.466} & 0.326 & \textbf{0.249} & \underline{\textit{0.420}} & 0.660 & \textbf{0.289} & \textbf{0.852} \\
        \textbf{Ours}$^{*}$ & Real & On. & 5.00k & 116.88k & \underline{\textit{0.546}} & \underline{\textit{1.433}} & 0.274 & \underline{\textit{0.214}} & 0.370 & 0.725 & \underline{\textit{0.272}} & / \\
        \rowcolor{gray!12}
        \multicolumn{3}{l}{\textit{Mean}} & / & / & 0.564 & 1.414 & 0.323 & 0.178 & 0.391 & 0.613 & 0.237 & 0.523 \\
        \bottomrule
    \end{tabular}
    \endgroup
\end{table*}

\subsection{Utterance Level}
We first examine a fundamental interactional question: to what extent do new patient expressions ground in the preceding dialogue, and how readily can their specific content be anticipated from context? This is investigated along two complementary dimensions: contextual continuity and conditional predictability.

\textbf{Contextual continuity.}
\emph{ECoh}~\cite{mendonca2024ecoh} measures the proportion of coherent patient turns relative to their preceding dialogue context, scored from the second patient turn onward and macro-averaged across consultations, where lower scores reflect weaker grounding in prior dialogue.

\textbf{Conditional predictability.}
\emph{BPB}, namely conditional bits per byte, measures the byte-normalized negative log-likelihood of patient text conditioned on the cumulative preceding dialogue history, capturing unprompted details and topic transitions that appear coherent yet remain hard to anticipate, with higher values indicating greater unpredictability. Specifically, we compute \emph{BPB} via Qwen2.5-0.5B and limit the cross-corpus comparison to Chinese datasets to control for language-dependent distributional variance.

\textbf{Findings.}
As shown in Table~\ref{tab:rq1-results}, authentic online consultations exhibit markedly weaker contextual continuity than offline and simulated encounters. Real online corpora yield a mean \emph{ECoh} of 0.564, which falls substantially below the 0.778 observed in simulated online datasets and the range of 0.694 to 0.865 in offline consultations. This confirms that patient contributions in real-world settings anchor far less consistently to the preceding dialogue context.
Conditional predictability reveals a sharper contrast. While the simulated dataset DISC-Med-SFT and the authentic corpus DiabetesCare display nearly identical \emph{ECoh} scores of 0.608 and 0.600, their \emph{BPB} values diverge drastically at 1.035 versus 1.466. This divergence confirms that contextual coherence masks substantial information uncertainty. More broadly, authentic online datasets consistently yield elevated \emph{BPB} ranging from 1.373 to 1.466, markedly exceeding the 1.035 to 1.272 recorded in simulated online dialogues and the baseline of 0.761 in offline clinical benchmarks. These findings show that authentic patient inputs are fundamentally under-constrained by prior discourse. Such unpredictability motivates treating incoming turns as partial observations for dynamic belief state inference. 

Crucially, by introducing unelicited clinical evidence, these unprompted expressions may alter physician action planning, which compels us to investigate how macro discourse progression is reshaped by such local state transitions.

\subsection{Dialogue Level}
We investigate dialogue-level progression through the interplay between patient-driven state deviations and physician agenda accommodation, examining whether authentic online consultations exhibit greater patient initiative in reshaping clinical trajectories alongside adaptive physician responses. To this end, we design five interaction metrics that quantify unsolicited information disclosure, patient-initiated agenda changes, and subsequent physician uptake through explicit behavioral criteria and dialogue-level aggregation. Detailed metric procedures are provided in Appendix~\ref{app:interaction-metrics}.

\subsubsection{Patient perspective.}
To quantify how patients actively intervene in the consultation, we assess informational proactivity through unsolicited disclosures~\cite{delpiccolo2007sequence} and conversational divergence through patient-initiated topic shifts~\cite{campion2004topic}.

\textbf{Informational proactivity.}
\emph{Discl.}, namely unsolicited clinical disclosures. Inspired by Del Piccolo et al.~\cite{delpiccolo2007sequence}, we design this metric to measure the proportion of valid patient turns volunteering unprompted facts beyond the scope of active physician inquiries, macro-averaged across consultations, where higher scores reflect more frequent evidence injection. Responses to open-ended invitations are not classified as unsolicited disclosures.

\textbf{Conversational divergence.}
\emph{Redir.}, namely unprompted conversational redirections. Inspired by Campion and Langdon~\cite{campion2004topic}, we design this metric to measure the proportion of patient turns that introduce a novel clinical concern relative to all valid patient turns, macro-averaged across consultations, capturing unelicited agenda transitions that diverge from the active inquiry, with higher values indicating more frequent structural disruptions. Specifically, we identify these shifts only when a patient initiates a distinct health focus rather than elaborating on current symptoms or responding to physician topic transitions.

\textbf{Aggregate autonomy.}
\emph{Any.}, namely overall patient-initiated actions. Inspired by Laukka et al.~\cite{laukka2020portals}, we design this metric to measure the proportion of patient turns that exhibit \emph{Discl.}, \emph{Redir.}, or both, macro-averaged across consultations, capturing the cumulative propensity of patients to intervene in the dialogue trajectory, with higher values indicating stronger conversational self-direction. Each turn is counted only once even if both behaviors co-occur.

\textbf{Findings.}
As shown in Table~\ref{tab:rq1-results}, authentic online consultations display a marked surge in patient-driven state deviations across both informational and agenda levels.

In terms of informational proactivity, authentic dialogues yield a mean \emph{Discl.} rate of 0.323, more than double the 0.149 observed in simulated interactions and substantially surpassing offline benchmarks ranging from 0.105 to 0.135. Crucially, every authentic online corpus exceeds all evaluated simulated online datasets on this measure, establishing that real-world patients routinely inject unprompted clinical evidence into the encounter.

Beyond volunteering information within ongoing inquiries, patients also frequently steer the conversational agenda itself. 
Authentic consultations exhibit a mean \emph{Redir.} rate of 0.178, over twice that of simulated dialogues at 0.076 and an order of magnitude above offline corpora at 0.018 to 0.025. Driven by both unprompted disclosures and conversational redirections, aggregate autonomy reaches an \emph{Any.} rate of 0.391 in authentic settings, roughly doubling simulated rates at 0.191 and nearly tripling offline interactions spanning 0.116 to 0.147. These disparities highlight that authentic online consultations operate under an evolving, patient-driven dynamic distinct from both offline and simulated baselines.

\subsubsection{Physician perspective.}
Although patients frequently steer the interaction through unprompted disclosures and shifts, whether physician behavior also exhibits distinctive online characteristics, such as simply following the patient's lead, remains less clear. We therefore investigate this dynamic through conditional responsiveness and the overall prevalence of physician uptake.

\textbf{Conditional responsiveness.}
\emph{Cond.}, namely conditional uptake of patient initiative. Inspired by Demszky et al.~\cite{demszky2021uptake}, we design this metric to measure the proportion of patient-initiated turns directly incorporated by the physician in the subsequent turn, macro-averaged across consultations, where higher scores reflect stronger immediate responsiveness to emergent patient inputs.

\textbf{Collaborative alignment.}
\emph{Colla.}, namely global conversational uptake. We design this metric to measure the proportion of valid patient-physician turn pairs across the entire consultation in which unprompted patient initiative is taken up by the physician, macro-averaged across dialogues. While conditional responsiveness reflects how reliably physicians respond once an initiative occurs, \emph{Colla.} captures the overall frequency of such reciprocal exchanges across the entire consultation, with higher values indicating more prevalent bilateral interactions.

\textbf{Findings.}
Authentic online consultations exhibit a lower \emph{Cond.} rate of 0.613 compared to 0.789 in simulated dialogues. This gap may reflect physicians' selective prioritization of diagnostic relevance when faced with frequent unprompted contributions. When evaluated through \emph{Colla.}, however, the pattern reverses. Authentic online consultations achieve a mean score of 0.237, exceeding simulated online consultations at 0.146 and offline corpora at 0.068 to 0.069.
This contrast indicates that while physicians respond selectively to individual patient initiatives, they frequently accommodate such contributions at the overall dialogue level. We speculate that this accommodative pattern may also be influenced by platform dynamics, where user ratings and satisfaction feedback implicitly incentivize physicians to preserve rapport.

Crucially, such a tension between selective uptake and flexible conversational steering exposes the limits of reflexive dialogue policies. It is therefore crucial to first evaluate the interactional context before formulating a response, identifying the necessary prerequisites that justify when to accommodate and when to steer.

\subsection{Corpus Level}
Finally, we examine whether the characteristics identified above are pervasive across authentic online datasets and substantially distinct from simulated and offline corpora.

\textbf{MAUVE.}
We evaluate the global text distribution of dialogue corpora using \emph{MAUVE}~\cite{pillutla2021mauve}, treating our authentic online dataset as the reference distribution, where higher \emph{MAUVE} scores reflect closer distributional similarity to genuine online clinical dialogues. Specifically, to eliminate language effects, we compute this metric exclusively on Chinese corpora.

\textbf{Findings. }
Authentic web corpora exhibit strong distributional cohesion. Using our corpus as the reference, independently collected real web datasets achieve a mean MAUVE of 0.523, substantially higher than simulated web corpora at 0.274 and the evaluated simulated offline corpus at 0.005, as reported in Table~\ref{tab:rq1-results}. Together with the preceding findings, this consistency further confirms that the distinctive linguistic and interactional challenges identified above are shared across authentic web consultations and are largely absent from simulated and offline settings.

\begin{figure*}[t]
    \centering
    \includegraphics[width=0.93\textwidth]{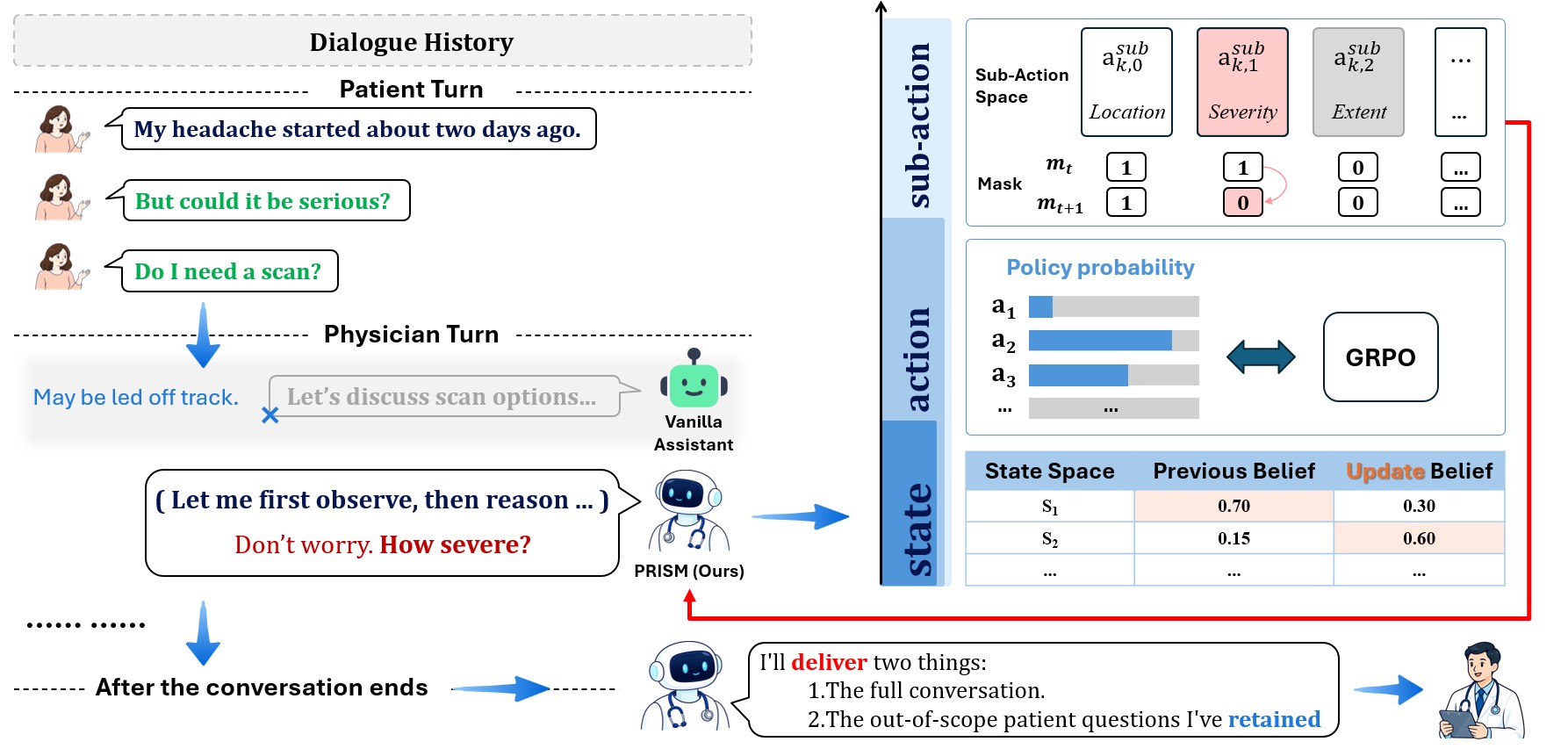}
    \caption{Overview of PRISM.}
    \Description{Overview of PRISM.}
    \label{fig:main}
\end{figure*}

\subsection{Summary: Addressing RQ1}
Our multi-tiered analysis provides a complementary evaluation across microscopic utterances, macroscopic discourse progression from both patient and physician perspectives, and global corpus distributions. Together, these results demonstrate that authentic online consultations are characterized by weakened contextual grounding, unprompted information injection, and an ongoing tension between selective diagnostic filtering and flexible conversational steering. Crucially, these traits are pervasive across authentic online corpora yet largely absent from simulated and offline settings.

This complex interactional environment exposes the limitations of reflexive dialogue policies. Consequently, it becomes essential to formulate online clinical consultation as an adaptive decision process under partial observability, dynamically inferring latent states to balance clinical efficacy against conversational rapport.

\section{PRISM}
Grounded in the intuition of partially observable decision processes, PRISM directly estimates the latent state posterior $b_\theta(s_t \mid h_t)$ conditioned on the observable history $h_t$.
\begin{equation}
p(s_t, \mathbf{a}_t, y_t | h_t) = b_\theta(s_t | h_t) \cdot \pi_\theta(\mathbf{a}_t | h_t, s_t) \cdot p_\psi(y_t | h_t, s_t, \mathbf{a}_t),
\end{equation}
where $b_\theta(s_t | h_t)$ performs \emph{belief estimation} by inferring the posterior distribution over candidate states $\mathcal{S}$, $\pi_\theta(\mathbf{a}_t | h_t, s_t)$ executes the \emph{policy decision} with its primary action drawn from $\mathcal{A}(s_t)$, and $p_\psi(y_t | h_t, s_t, \mathbf{a}_t)$ conducts \emph{response realization} to render the chosen decision into a patient-facing utterance $y_t$. Here, $\mathbf{a}_t$ comprises a primary action $a_t$ and, when applicable, a sub-action $a_{\mathrm{sub},t}$.

\subsection{Belief Estimation}
In authentic consultations, fragmented and evolving patient inputs cannot be reliably mapped to a deterministic trajectory, so PRISM maintains a posterior belief distribution over plausible states to preserve unresolved clinical possibilities.

Derived from authentic hospital consultation workflows, we maintain a set of triage states $\mathcal{S}_{\text{triage}}$. In addition, we incorporate two control states. \texttt{ADVANCE} indicates that information collection required by the current triage state is satisfied, masking its weight to zero in subsequent turns to retire it from the active agenda and lower decision complexity. In contrast, \texttt{DELIVER} indicates that the model judges the case-specific information collection to be sufficient and terminates automated triage for physician handover. 

Formally, we define the complete state space as $\mathcal{S} = \mathcal{S}_{\text{triage}} \cup \{\texttt{ADVANCE}, \texttt{DELIVER}\}$. PRISM tracks the conversational ambiguity arising from fragmented patient disclosures by maintaining a posterior belief distribution $\hat{b}_t$ over all candidate states in $\mathcal{S}$.

Specifically, given the state-prediction context $c_t^s = h_t \oplus \text{\texttt{<STATE>}}$, PRISM scores each candidate state $s = (s_1, \dots, s_{L_s}) \in \mathcal{S}$ via its sequence log-likelihood:
\begin{equation}
    \ell_{\theta}(s; h_t) = \sum_{j=1}^{L_s} \log p_{\theta}(s_j \mid c_t^s, s_{<j}).
\end{equation}
To prune resolved clinical requirements, we maintain a binary availability mask $m_t(s) \in \{0, 1\}$, where $m_t(s) = 0$ if state $s$ has been satisfied by \texttt{ADVANCE} prior to turn $t$, and $m_t(s) = 1$ otherwise. Applying this mask yields the normalized belief distribution:
\begin{equation}
    \hat{b}_t(s) = \frac{m_t(s) \exp \ell_{\theta}(s; h_t)}{\sum_{s' \in \mathcal{S}} m_t(s') \exp \ell_{\theta}(s'; h_t)}, \quad \forall s \in \mathcal{S}.
\end{equation}
This formulation allows incoming unprompted evidence in $u_t$ to continuously reweight the belief distribution over currently available states, maintaining awareness of alternative clinical trajectories even when prior inquiries remain pending.

To interface with the downstream policy while grounding decisions in the estimated belief, PRISM extracts the maximum a posteriori (MAP) state:
\begin{equation}
    \hat{s}_t = \arg\max_{s \in \mathcal{S}} \hat{b}_t(s).
\end{equation}
This predicted state is serialized into the discrete textual span $z_t^s = \text{\texttt{<STATE>}} \hat{s}_t \text{\texttt{</STATE>}}$, which serves as the explicit structural interface to constrain subsequent action selection. When $\hat{s}_t=\texttt{DELIVER}$, PRISM terminates automated triage and hands over the collected clinical information and recorded patient concerns to the physician, without selecting another inquiry action.

\subsection{Policy Decision}
In clinical triage, information requirements vary across stages, making the candidate action space strictly state-dependent. While symptom elicitation spans physiological systems such as the respiratory, circulatory, and digestive domains, family history inquiries target only kinship and familial health, thereby restricting the permissible inquiry directions to the inferred clinical state.

Yet, merely restricting the inquiry to the admissible action space $\mathcal{A}(s)$ remains insufficient for rigorous clinical assessment. A single inquiry action $a \in \mathcal{A}(s)$, such as the disease timeline, still entails multiple essential clinical factors including the time of first occurrence, duration, progression speed, and evolving trends. To capture these multi-faceted requirements, we formulate an expanded sub-action space $\mathcal{A}_{\mathrm{sub}}(a)$ conditioned on the selected action $a$. Each sub-action space also includes \texttt{EXIT}.

In practical use, following the identification of the MAP state $\hat{s}_t$, the structural delimiter \texttt{<ACTION>} is appended to the predicted state span $z_t^s = \text{\texttt{<STATE>}} \hat{s}_t \text{\texttt{</STATE>}}$ to establish the conditioned decision interface. Letting $\widetilde{\mathcal{A}}_t = \operatorname{ser}(\mathcal{A}(\hat{s}_t))$ denote the serialized representation of candidate actions under $\hat{s}_t$, the policy context is formulated as
\begin{equation}
    c_t^a = h_t \oplus z_t^s \oplus \widetilde{\mathcal{A}}_t \oplus \text{\texttt{<ACTION>}}.
\end{equation}
Each action set $\mathcal{A}(s)$ incorporates the dedicated terminal action \texttt{EXIT}. Predicting \texttt{EXIT} signifies that information collection for the current stage is sufficient, which triggers \texttt{ADVANCE} in the state tracker to retire this state from the active agenda. Conditioned on the decision context $c_t^a$, the policy decomposes the inquiry generation into two successive steps, first selecting the high-level action
\begin{equation}
    \hat{a}_t = \arg\max_{a \in \mathcal{A}(\hat{s}_t)} \pi_{\theta}(a \mid c_t^a).
\end{equation}

If a non-terminal action $\hat{a}_t \neq \text{\texttt{EXIT}}$ is selected, PRISM refines the inquiry by selecting a target from the sub-action space $\mathcal{A}_{\mathrm{sub}}(\hat{a}_t)$. To prevent redundant questioning on explored clinical dimensions, we restrict candidate targets via a dynamic tracking mask. Specifically, letting $m_t^{\mathrm{sub}}(a_{\mathrm{sub}}; a) \in \{0, 1\}$ indicate whether sub-action $a_{\mathrm{sub}}$ under action $a$ remains unqueried, with $m_t^{\mathrm{sub}}(\text{\texttt{EXIT}}; a)=1$ throughout, the active sub-action space at turn $t$ is defined as
\begin{equation}
    \mathcal{A}_{\mathrm{sub},t}(\hat{a}_t) = \left\{ a_{\mathrm{sub}} \in \mathcal{A}_{\mathrm{sub}}(\hat{a}_t) \;\middle|\; m_t^{\mathrm{sub}}(a_{\mathrm{sub}}; \hat{a}_t) = 1 \right\}.
\end{equation}

Letting $\widetilde{\mathcal{A}}_{\mathrm{sub},t} = \operatorname{ser}(\mathcal{A}_{\mathrm{sub},t}(\hat{a}_t))$ denote the serialized available sub-actions, the sub-action context is formed by appending the chosen primary action and the unmasked candidates: $c_t^{a_{\mathrm{sub}}} = c_t^a \oplus \hat{a}_t \oplus \widetilde{\mathcal{A}}_{\mathrm{sub},t}$. The policy then resolves the concrete inquiry target:
\begin{equation}
    \hat{a}_{\mathrm{sub},t} = \arg\max_{a_{\mathrm{sub}} \in \mathcal{A}_{\mathrm{sub},t}(\hat{a}_t)} \pi_{\theta}(a_{\mathrm{sub}} \mid c_t^{a_{\mathrm{sub}}}).
\end{equation}
Upon selection, PRISM updates $m_{t+1}^{\mathrm{sub}}(\hat{a}_{\mathrm{sub},t}; \hat{a}_t)=0$ only for non-terminal sub-actions. The parent action $\hat{a}_t$ remains unmasked, allowing later turns to return to its remaining sub-actions until the state transitions. The selected decisions are serialized into the structural action span:
\begin{equation}
    z_t^a = 
    \begin{cases} 
        \text{\texttt{<ACTION>}} \oplus \hat{a}_t \oplus \text{\texttt{</ACTION>}}, & \text{if } \hat{a}_t = \text{\texttt{EXIT}}, \\ 
        \text{\texttt{<ACTION>}} \oplus \operatorname{ser}(\hat{a}_t, \hat{a}_{\mathrm{sub},t}) \oplus \text{\texttt{</ACTION>}}, & \text{otherwise}, 
    \end{cases}
\end{equation}
supplying the explicit directive for subsequent response realization.

To optimize decision-making across both levels, PRISM employs Group Relative Policy Optimization (GRPO)~\cite{shao2024deepseekmath}. For either decision level $q \in \{a, a_{\mathrm{sub}}\}$, a group of $G$ candidate completions $O_t^q = \{o_i\}_{i=1}^G$ is sampled from the old policy $\pi_{\theta_{\text{old}}}(\cdot \mid c_t^q)$. Primary action candidates are drawn from $\mathcal{A}(\hat{s}_t)$ (including \texttt{EXIT}), whereas sub-action candidates are drawn from the unmasked space $\mathcal{A}_{\mathrm{sub},t}(\hat{a}_t)$. Each candidate is evaluated against the annotated reference $q_t^*$ via a level-specific reward function $R_q$, yielding the group-relative advantage:
\begin{equation}
    r_i = R_q(o_i, q_t^*), \qquad \hat{A}_i = \frac{r_i - \operatorname{mean}(\{r_k\}_{k=1}^G)}{\operatorname{std}(\{r_k\}_{k=1}^G)}.
\end{equation}
For the $j$-th token in completion $o_i$, the importance sampling probability ratio is defined as:
\begin{equation}
    \rho_{i,j}(\theta) = \frac{\pi_{\theta}(o_{i,j} \mid c_t^q, o_{i,<j})}{\pi_{\theta_{\text{old}}}(o_{i,j} \mid c_t^q, o_{i,<j})}.
\end{equation}
The policy parameters $\theta$ are updated by maximizing the token-level clipped objective:
\begin{equation}
\begin{aligned}
\mathcal{J}_{\text{GRPO}}^q(\theta)
&= \mathbb{E}_{O_t^q} \Bigg[ \frac{1}{G} \sum_{i=1}^G
\frac{1}{|o_i|} \sum_{j=1}^{|o_i|} \\
&\quad \Big( \min \big[ \rho_{i,j}(\theta) \hat{A}_i,
\bar{\rho}_{i,j}(\theta) \hat{A}_i \big]
-\beta \mathcal{K}_{i,j} \Big) \Bigg],
\end{aligned}
\end{equation}
where $\bar{\rho}_{i,j}(\theta) = \operatorname{clip}(\rho_{i,j}(\theta), 1-\epsilon, 1+\epsilon)$, and $\mathcal{K}_{i,j} = \mathbb{D}_{\text{KL}}\big(\pi_{\theta}(\cdot \mid c_t^q, o_{i,<j}) \parallel \pi_{\text{ref}}(\cdot \mid c_t^q, o_{i,<j})\big)$ denotes the token-level KL divergence penalty against the reference model $\pi_{\text{ref}}$, $\epsilon$ is the clipping parameter, and $\beta$ controls regularization strength. This formulation enables PRISM to learn high-level inquiry targeting, fine-grained sub-action selection, and appropriate stage termination.

\begin{table*}[t]
    \centering
    \caption{Next-question generation performance of PRISM and baseline systems.}
    \label{tab:agent-response-main}
    \begingroup
    \begin{tabular}{lrrrrrrrrrr}
        \toprule
        Model & EM & ROUGE-1 & ROUGE-2 & ROUGE-L & BLEU-4 & chrF & Sem. & BertScore$_{P}$ & BertScore$_{R}$ & BertScore$_{F1}$ \\
        \midrule
        Qwen3.8-27B & 0.00 & 22.07 & 8.53 & 18.88 & 3.07 & 8.52 & 58.96 & 61.97 & 73.44 & 67.12 \\
        Qwen3.8-Max & 0.00 & 21.55 & 8.53 & 18.54 & 3.28 & 8.87 & 59.03 & 61.35 & 73.30 & 66.71 \\
        DeepSeek-V4-Pro & 0.00 & 22.76 & 9.46 & 19.72 & 3.48 & 9.32 & 59.01 & 61.79 & 73.63 & 67.10 \\
        MiniMax-M3 & 0.00 & 22.03 & 9.37 & 19.27 & 3.40 & 9.38 & 59.49 & 60.65 & 73.17 & 66.23 \\
        GPT-6-Astra & 0.04 & 21.00 & 8.57 & 17.96 & 2.93 & 8.59 & 58.62 & 60.06 & 72.63 & 65.65 \\
        \midrule
        Codex & 0.00 & 23.30 & 8.92 & 19.85 & 3.61 & 9.21 & 60.05 & 62.43 & 73.71 & 67.53 \\
        MiMoCode & \underline{\textit{0.18}} & \underline{\textit{31.30}} & \underline{\textit{13.23}} & \underline{\textit{28.94}} & \underline{\textit{7.65}} & \underline{\textit{10.10}} & \underline{\textit{60.59}} & \underline{\textit{71.31}} & \underline{\textit{75.27}} & \underline{\textit{73.14}} \\
        \midrule
        Doctor-R1 & 0.00 & 23.41 & 9.49 & 20.67 & 3.57 & 8.78 & 58.40 & 62.98 & 73.06 & 67.53 \\
        MeDxAgent & 0.00 & 27.67 & 10.78 & 25.28 & 5.26 & 8.85 & 58.76 & 69.13 & 74.91 & 71.80 \\
        \midrule
        Qwen3-8B-SFT & 0.00 & 23.97 & 9.68 & 21.16 & 3.75 & 8.97 & 58.59 & 63.53 & 73.31 & 67.96 \\
        PRISM & \textbf{3.90} & \textbf{43.57} & \textbf{26.02} & \textbf{42.05} & \textbf{18.43} & \textbf{16.70} & \textbf{65.04} & \textbf{79.72} & \textbf{77.78} & \textbf{78.65} \\
        \bottomrule
    \end{tabular}
    \endgroup
\end{table*}

\subsection{Response Realization}
Finally, conditioned on the inferred MAP state and the selected clinical action, PRISM synthesizes the patient-facing utterance $y_t$. We concatenate the dialogue history $h_t$ with the inferred control spans into the realization context $c_t^y = h_t \oplus z_t^s \oplus z_t^a$. This conditioning explicitly decouples high-level clinical decision-making from surface realization, ensuring that communicative goals remain strictly grounded in the structured decisions.

Given the context $c_t^y$, the response $y_t$ is decoded autoregressively under parameters $\psi$. We optimize the realization module using standard cross-entropy loss with teacher forcing on ground-truth utterances $y_t^* \in \mathcal{D}$, where $K=|y_t^*|$:
\begin{equation}
\mathcal{L}_{\mathrm{response}} = - \sum_{k=1}^{K} \log p_{\psi}\!\left(y_{t,k}^{*} \mid c_t^y, y_{t,<k}^{*}\right).
\end{equation}
During deployment, the generated control sequence directly steers the clinical tone and query phrasing, preventing latent policy drift in free-form generation.

\section{Experiments}

\subsection{Dataset}
Our evaluation corpus consists of authentic online medical dialogues collected from a major telemedicine platform. Grounded in the clinical triage protocol of a leading tertiary-care teaching hospital, we annotate each conversational turn with its triage state $s \in \mathcal{S}$, inquiry action $a \in \mathcal{A}(s)$, and specific clinical facet $a_{\mathrm{sub}} \in \mathcal{A}_{\mathrm{sub}}(a)$.

In total, we curate 450 dialogues across nine distinct disease categories, allocating 50 cases to each condition and yielding an average of 1400 utterances per category. 
Dataset selection jointly considers patient gender, age, and consultation timing, with comprehensive distributional statistics detailed in Appendix~\ref{app:case-distribution}.

\subsection{Privacy Compliance}
Confidentiality-wise, the annotation pipeline is powered by an open-source DeepSeek-V4-Pro model deployed on three private internal nodes, each equipped with $8 \times \text{H20 140GB}$ GPUs. To enforce stringent annotation quality, each sample undergoes three independent annotation passes, and we retain only instances where all three passes reach complete agreement. Constraints on external clinical evaluation are discussed in Appendix~\ref{app:limitations}.

Furthermore, to enable comparative evaluation against commercial API baselines without compromising privacy, we employ the local model to sanitize the corpus. This sanitization protocol redacts private clinical disclosures, paraphrases semi-sensitive identifiers such as patient names and locations via synonymous rewriting, and selectively restructures pertinent utterance phrasing while preserving conversational coherence and clinical validity.

\subsection{Baselines}
We evaluate PRISM on response generation against foundation model baselines, models accessed via commercial APIs, agent baselines, and related methods. We also evaluate state-guided action selection and conduct ablation studies to examine the contribution of explicit consultation-state information to response generation. Implementation and training details of PRISM are provided in Appendix~\ref{app:implementation-training}.

Our foundation model baseline is Qwen3-8B-SFT, fine-tuned directly on the consultation dataset without explicit latent state modeling as an end-to-end baseline. We evaluate Qwen3.8-Max, DeepSeek-V4-Pro, MiniMax-M3, and GPT-6-Astra-xhigh through commercial APIs, using the highest available reasoning and thinking budget where supported. Agent baselines comprise two general-purpose commercial systems evaluated on question generation from consultation contexts: Codex with GPT-6-Astra-xhigh and MiMoCode with MiMo-V2.6-Pro. We further compare against Doctor-R1~\cite{lai2025doctorr1} and MeDxAgent~\cite{sanghvi2026medxagent}, two related methods for interactive clinical consultation. Doctor-R1 uses its Qwen3-8B-based checkpoint, while MeDxAgent is implemented with Qwen3.8-27B; both are adapted to our triage task.

\textbf{Metrics.}
For action-policy evaluation, we report accuracy, precision, recall, and F1 against annotated clinical decisions. We evaluate generated questions against references across exact, lexical, and semantic dimensions. We report EM for exact matching, character-level ROUGE-1/2/L F1, BLEU-4 via SacreBLEU with Chinese tokenization, and chrF ($\beta=2$) for lexical overlap, alongside multilingual MiniLM cosine similarity (Sem.) and BERTScore P/R/F1 for semantic alignment. All scores are reported on a 100-point scale, with higher values indicating higher fidelity.

\subsection{Response Generation Analysis}
As Table~\ref{tab:agent-response-main} shows, PRISM dominates all ten metrics across foundation models, commercial agents, and specialized medical baselines, establishing a consistent advantage at both the lexical and semantic levels. Agreement across exact, lexical, character-sequence, and embedding metrics supports a benefit beyond one particular overlap measure.

On one hand, we observe a distinct advantage at the lexical level. Higher lexical scores stem from greater vocabulary overlap, which in our case likely reflects more frequent matches in clinical terminology. Compared to MiMoCode, PRISM improves ROUGE-1 from 31.30 to 43.57, nearly doubles ROUGE-2 from 13.23 to 26.02, and more than doubles BLEU-4 from 7.65 to 18.43. We attribute this advantage to PRISM's state-conditioned action selection, which provides concrete inquiry targets for response generation. Furthermore, even under the stringent Exact Match metric where most baselines remain near 0.00, PRISM achieves 3.90, confirming instances of exact agreement with reference inquiries.

On the other hand, PRISM also achieves the highest semantic similarity, advancing Sem. from 60.59 to 65.04 alongside a marked BertScore Precision increase from 71.31 to 79.72. Taken together, the lexical and semantic gains support the effectiveness of our state-guided action policy in directing both the content and phrasing of clinical inquiries.

\subsection{Action Policy Analysis: Addressing RQ2}
We evaluate PRISM's state-guided action policy against annotated clinical decisions in Table~\ref{tab:decision-results}. PRISM achieves an accuracy of 76.39\%, surpassing the strongest foundation baseline MiniMax-M3 by 22.40 percentage points. Its F1 score reaches 74.72\%, outperforming the highest baseline score of 51.82\% from DeepSeek-V4-Pro by 22.90 points. These substantial margins demonstrate the effectiveness of state-guided action selection in identifying the next clinical steps.

Meanwhile, commercial agents such as Codex and MiMoCode achieve accuracies of only 52.78\% and 52.90\%, plateauing near general foundation models. This indicates that general code or interactive reasoning abilities offer limited transfer to clinical decision-making. We attribute PRISM's advantage to explicit belief estimation and state-conditioned policy optimization, which ground action selection in the inferred consultation state and its unresolved information requirements. The resulting action policy supplies concrete inquiry targets for subsequent response generation.

\begin{table}[t]
    \centering
    \caption{Clinical decision performance.}
    \label{tab:decision-results}
    \begingroup
    \begin{tabular}{lcccc}
        \toprule
        Model & Acc.$\uparrow$ & Recall$\uparrow$ & Precision$\uparrow$ & F1$\uparrow$ \\
        \midrule
        Qwen3-8B-SFT & 43.07 & 43.07 & 47.88 & 40.95 \\
        Qwen3.8-27B & 46.19 & 46.19 & 51.70 & 44.40 \\
        Qwen3.8-Max & 51.76 & 51.76 & 60.68 & 48.31 \\
        DeepSeek-V4-Pro & 53.65 & 53.65 & 60.18 & \underline{\textit{51.82}} \\
        MiniMax-M3 & \underline{\textit{53.99}} & \underline{\textit{53.99}} & 60.44 & 51.68 \\
        GPT-6-Astra & 50.98 & 50.98 & 58.58 & 48.24 \\
        \midrule
        Codex & 52.78 & 52.78 & \underline{\textit{61.68}} & 49.65 \\
        MiMoCode & 52.90 & 52.90 & 58.34 & 50.25 \\
        \midrule
        PRISM & \textbf{76.39} & \textbf{76.39} & \textbf{75.88} & \textbf{74.72} \\
        \bottomrule
    \end{tabular}
    \endgroup
\end{table}

\begin{figure}[ht]
    \centering
    \includegraphics[width=\columnwidth]{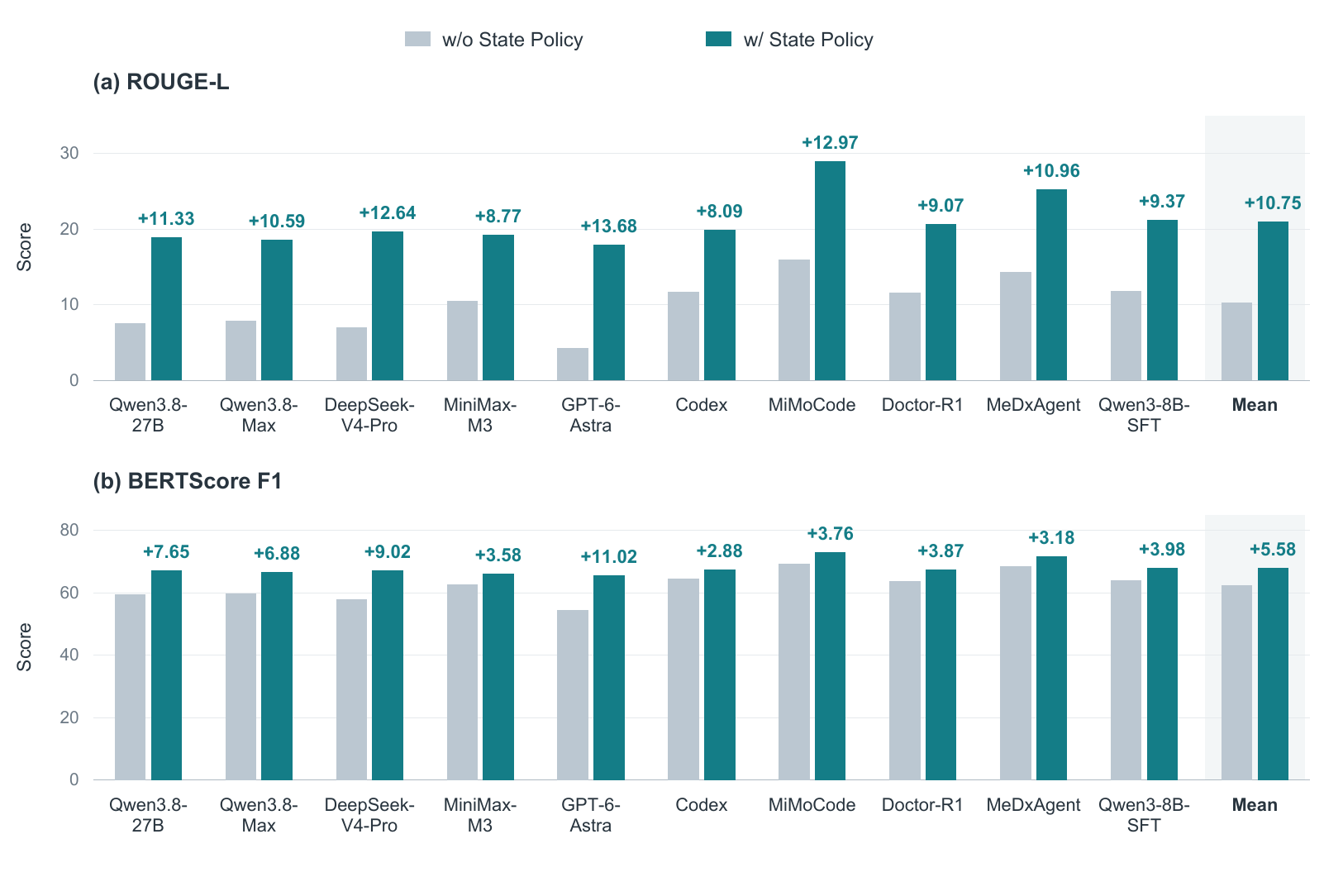}
    \caption{Response generation with and without oracle state policy guidance. Labels indicate score gains.}
    \Description{Response generation with and without oracle state policy guidance. Labels indicate score gains.}
    \label{fig:policy-ablation}
\end{figure}
\subsection{State Policy Ablation: Addressing RQ3}
To isolate the contribution of consultation-state grounding, we provide each baseline with a unified oracle state policy supplying only the ground-truth consultation state under identical dialogue histories. Each model determines its action policy through its own chain-of-thought reasoning and subsequently generates the response. As shown in Figure~\ref{fig:policy-ablation}, this state conditioning consistently improves all ten models across both ROUGE-L and BERTScore F1, yielding average gains of 10.75 and 5.58 points, respectively. These gains extend to advanced commercial agents and specialized medical systems. MiMoCode improves by 12.97 points in ROUGE-L and 3.76 points in BERTScore F1, while MeDxAgent gains 10.96 and 3.18 points. 

Crucially, these consistent improvements show that even frontier models benefit substantially from explicit state grounding while retaining responsibility for action planning. Strong chain-of-thought reasoning alone does not remove the need to identify the current consultation state before deciding how to proceed.

\section{Related Work}

Medical dialogue systems address consultation through three mechanisms: medical response alignment, coordinated clinical reasoning, and interactive information acquisition. Response alignment adapts LLMs to patient inquiries using real exchanges in ChatDoctor~\cite{li2023chatdoctor}, physician and distilled responses in HuatuoGPT~\cite{zhang2023huatuogpt}, and reconstructed consultations in DISC-MedLLM~\cite{discmed}. Coordinated reasoning distributes clinical problems across expert roles through prompted deliberation in MedAgents~\cite{tang2024medagents} and reinforcement learning in MMedAgent-RL~\cite{xia2026mmedagent}, primarily strengthening how systems reason over available evidence. Interactive approaches additionally acquire evidence through self-play-trained diagnostic dialogue in AMIE~\cite{tu2025towards} and staged diagnostic procedures in ClinDiag-GPT~\cite{chen2026grounding}. However, their interaction mechanisms are developed extensively through constructed encounters, including simulated history taking in MEDDxAgent~\cite{rose2025meddxagent}, virtual clinical experience in Agent Hospital~\cite{li2025agenthospital}, and simulated consultations in AgentClinic~\cite{schmidgall2026agentclinic}. Across these approaches, the central emphasis remains on improving response quality, coordinating internal expertise, or eliciting evidence toward a predefined clinical objective. A fundamental challenge arises, however, when subsequent patient disclosures reshape which concerns require attention and what must be clarified next. PRISM addresses this challenge by explicitly maintaining beliefs over evolving consultation states and conditioning sequential action selection on the inferred MAP state under partial observability.

\section{Conclusion}
This paper tackles the irregular state transitions of online medical consultation, where unprompted patient disclosures disrupt traditional inquiry workflows. To resolve this partial observability, we propose PRISM, a framework that factorizes clinical triage into latent belief tracking, state-constrained policy optimization, and response realization. Across extensive benchmarks, PRISM decisively outperforms frontier LLMs and fine-tuned baselines in both action selection and response generation, confirming that explicit state grounding is vital for robust clinical triage.

\clearpage
\bibliographystyle{ACM-Reference-Format}
\bibliography{sample-base}

\clearpage
\appendix
\setlength{\emergencystretch}{2em}
\flushbottom
\makeatletter\@ACM@balancefalse\makeatother
\renewcommand{\thetable}{\Alph{section}\arabic{table}}
\renewcommand{\theHtable}{\Alph{section}.\arabic{table}}
\setcounter{table}{0}
\captionsetup[table]{position=top,skip=6pt}
\definecolor{prismtrain}{gray}{0.28}
\newcommand{\prismexample}[2]{%
\begin{mdframed}[linecolor=prismtrain,linewidth=2pt,backgroundcolor=black!6,
leftline=true,rightline=false,topline=false,bottomline=false,
innerleftmargin=4pt,innerrightmargin=3pt,innertopmargin=1pt,innerbottommargin=2pt,
skipabove=2pt,skipbelow=2pt,splittopskip=2pt,splitbottomskip=2pt,
frametitle={#1},frametitlefont=\small\bfseries\color{black},frametitlebackgroundcolor=black!16,
frametitleaboveskip=2pt,frametitlebelowskip=1pt,repeatframetitle=true]
\small\setlength{\baselineskip}{9pt}\setlength{\parindent}{0pt}\setlength{\parskip}{0pt}#2
\end{mdframed}}

\section{Distribution of the 450 Evaluation Cases}
\label{app:case-distribution}
The evaluation set contains 450 consultation dialogues, with 50 cases in each of nine categories: type~1 diabetes, endocrine disorders, prostatitis, genital warts, chronic obstructive pulmonary disease, cosmetic laser treatment, lung cancer, lumbar disc herniation, and glaucoma. Alongside balanced case categories, it covers gender, age, and consultation month; Table~\ref{tab:app-demographics} reports these demographic and temporal distributions.

\par\addvspace{6pt}
\noindent\begin{minipage}{\linewidth}
\definecolor{genderbg}{RGB}{225,237,249}
\definecolor{agebg}{RGB}{230,242,226}
\definecolor{monthbg}{RGB}{252,237,215}
\captionof{table}{Distribution of evaluation cases. Background colors distinguish gender (light blue), age (light green), and consultation month (light orange).}
\label{tab:app-demographics}
\centering\small\setlength{\tabcolsep}{3pt}
\begin{tabular}{@{}
>{\raggedright\arraybackslash}p{\dimexpr.27\linewidth-\tabcolsep\relax}
>{\raggedleft\arraybackslash}p{\dimexpr.11\linewidth-2\tabcolsep\relax}
>{\raggedleft\arraybackslash}p{\dimexpr.15\linewidth-2\tabcolsep\relax}
>{\raggedright\arraybackslash}p{\dimexpr.21\linewidth-2\tabcolsep\relax}
>{\raggedleft\arraybackslash}p{\dimexpr.11\linewidth-2\tabcolsep\relax}
>{\raggedleft\arraybackslash}p{\dimexpr.15\linewidth-\tabcolsep\relax}@{}}
\toprule
Characteristic & $n$ & \% & Month & $n$ & \% \\
\midrule
\cellcolor{genderbg}Female & \cellcolor{genderbg}250 & \cellcolor{genderbg}55.56 & \cellcolor{monthbg}Jan. & \cellcolor{monthbg}44 & \cellcolor{monthbg}9.78 \\
\cellcolor{genderbg}Male & \cellcolor{genderbg}200 & \cellcolor{genderbg}44.44 & \cellcolor{monthbg}Feb. & \cellcolor{monthbg}51 & \cellcolor{monthbg}11.33 \\
\cellcolor{agebg}Age $<18$ & \cellcolor{agebg}56 & \cellcolor{agebg}12.44 & \cellcolor{monthbg}Mar. & \cellcolor{monthbg}75 & \cellcolor{monthbg}16.67 \\
\cellcolor{agebg}18--29 & \cellcolor{agebg}96 & \cellcolor{agebg}21.33 & \cellcolor{monthbg}Apr. & \cellcolor{monthbg}75 & \cellcolor{monthbg}16.67 \\
\cellcolor{agebg}30--44 & \cellcolor{agebg}150 & \cellcolor{agebg}33.33 & \cellcolor{monthbg}May & \cellcolor{monthbg}48 & \cellcolor{monthbg}10.67 \\
\cellcolor{agebg}45--59 & \cellcolor{agebg}77 & \cellcolor{agebg}17.11 & \cellcolor{monthbg}Jun. & \cellcolor{monthbg}49 & \cellcolor{monthbg}10.89 \\
\cellcolor{agebg}60--74 & \cellcolor{agebg}62 & \cellcolor{agebg}13.78 & \cellcolor{monthbg}Jul. & \cellcolor{monthbg}47 & \cellcolor{monthbg}10.44 \\
\cellcolor{agebg}$\geq75$ & \cellcolor{agebg}9 & \cellcolor{agebg}2.00 & \cellcolor{monthbg}Aug. & \cellcolor{monthbg}50 & \cellcolor{monthbg}11.11 \\
--- & --- & --- & \cellcolor{monthbg}Sep. & \cellcolor{monthbg}11 & \cellcolor{monthbg}2.44 \\
\bottomrule
\end{tabular}
\end{minipage}\par\smallskip

\section{PRISM Implementation and Training Details}
\label{app:implementation-training}
\setcounter{table}{0}
\subsection{Clinical Schema Excerpt}
\label{app:schema}
The schema defines the hierarchy of triage states $\mathcal S_{\mathrm{triage}}$, primary actions $\mathcal A(s)$, and sub-actions $\mathcal A_{\mathrm{sub}}(a)$. Commercial confidentiality and privacy restrict disclosure of our partner's complete internal information-collection workflow. Table~\ref{tab:app-schema} provides representative entries to explain the structure and semantics and support adaptation to other tasks.

\par\addvspace{6pt}
\noindent\begin{minipage}{\linewidth}
\captionof{table}{A partial illustration of the clinical schema.}
\label{tab:app-schema}
\centering\small\setlength{\tabcolsep}{3pt}
\begin{tabular}{@{}>{\raggedright\arraybackslash}p{.23\linewidth}>{\raggedright\arraybackslash}p{.25\linewidth}>{\raggedright\arraybackslash}p{.45\linewidth}@{}}
\toprule
State & Primary action & Sub-action examples \\
\midrule
History of present illness & Time course & Onset; duration; progression \\
History of present illness & Related factors & Triggers; aggravation; relief \\
Past history & Allergy history & Allergen; reaction; severity \\
Personal / lifestyle & Sleep / routine & Bedtime; waking time; duration \\
Existing objective data & Existing tests & Name; date; value; report text \\
Notes / concerns & Visit goals & Diagnostic concerns; symptom relief \\
\bottomrule
\end{tabular}
\end{minipage}\par\smallskip

\subsection{Stage-Specific Training Procedures}
PRISM training centers on three components: belief estimation, decision policy learning, and response realization. The training, validation, and intermediate-pool counts reported below distinguish among question contexts, level-specific decision contexts, and cases; they differ from the 5,000 dialogues used for analysis in Section~3 and the 450 dialogues used for evaluation in Section~5.

\subsubsection{Belief Estimation}
Belief estimation conditions subsequent action selection. Training/validation contain 12,917/642 question contexts from 2,365/124 cases. History $h_t$ forms the main-text context $c_t^s$; the target is the token sequence of reference state $s_t^*$:
\begin{equation}
\mathcal L_s(\theta)=-\mathbb E_{(h_t,s_t^*)}\sum_{j=1}^{L_{s_t^*}}
\log p_\theta(s_j^*\mid c_t^s,s_{<j}^*).
\label{eq:app-state-loss}
\end{equation}
Here $s_j^*$ is target token $j$ and $s_{<j}^*$ its preceding tokens. Loss excludes inputs and padding. At inference, candidate sequence scores are masked and normalized as in Section~4 to obtain the policy's state belief.
\prismexample{State-training example.}{%
\textbf{Input $h_t$ (excerpt):} Doctor: ``Any discomfort?'' Patient: ``I have been constipated since my kidney biopsy; laxatives have not helped.''\\
\textbf{Candidate states (excerpt):} [History of present illness (HPI), past history, existing objective data, \ldots, \texttt{ADVANCE}, \texttt{DELIVER}].\\
\textbf{Target state $s_t^*$:} HPI.}

\subsubsection{Training Strategy for Policy Decision}
Six steps comprise \textbf{(I)} structured-label annotation, \textbf{(II)} teacher annotation, \textbf{(III)} evidence maker training, \textbf{(IV)} expansion, \textbf{(V)} fresh-backbone SFT, and \textbf{(VI)} final-policy GRPO. Steps~1--4 construct supervision; Steps~5--6 train the deployed policy.

For primary-action ($q=a$) or sub-action ($q=a_{\mathrm{sub}}$) selection, $c_t^q$ contains history, admissible choices, and annotated parent controls: state for primary actions, state plus primary action for sub-actions. The reference label is $q_t^*$; the rationale is $e$.

\textbf{Stage I: Structured-label annotation.}
Using privately deployed DeepSeek-V4-Pro (Section~5.2), we annotate each context's state $s_t^*$, primary action $a_t^*$, and, when applicable, sub-action $a_{\mathrm{sub},t}^*$. Each sample undergoes three passes; only samples with complete label agreement are retained as references for subsequent supervision and rewards.

\textbf{Stage II: Teacher annotation.}
We build rationale-bearing demonstrations for the maker, balancing primary/sub-action types while considering clinical categories and history length. We use GPT-6-Astra-xhigh as teacher: given $(c_t^q,q_t^*)$, it explains the reference choice with a rationale $e$ grounded only in visible input evidence.
\prismexample{Teacher-annotation example: primary action.}{%
\textbf{Task level $q$:} $a$ (primary-action selection).\\
\textbf{Teacher input $c_t^a$:} State: HPI. History: red rash was noticed on a child's palms after school; duration is recorded as ``within one week.'' Candidate actions: [time course, characteristics, current medication, \ldots, \texttt{EXIT}].\\
\textbf{Reference supplied to teacher $a_t^*$:} Time course.\\
\textbf{Teacher output:} Rationale $e$: ``Duration is only known to be within one week; exact onset and subsequent changes remain unclear, so ask about the time course.'' Target $a_t^*$: time course.}
We use Kimi K3 to retain rationales supported by the input, consistent with the reference label, and within information collection.

\textbf{Stage III: Evidence Maker Training.}
We initialize $\pi_\phi$ from Qwen3-8B $M_0$ and train on 1,250 primary-action and 1,750 sub-action contexts to generate rationale--choice pairs. Two reviewed demonstrations $F_t$ match the decision level and approximate history length, preferably one same-category and one cross-category case. Retrieval excludes the current reference label; training excludes seed cases.
\prismexample{Maker-input example: primary action.}{%
\textbf{Task:} Produce a rationale and the next primary action for the current case.\\
\textbf{Demonstrations $F_t$:} Both have state HPI and candidates [time course, characteristics, current medication, \ldots, \texttt{EXIT}].\\
\textbf{Example 1:} History: red rash was noticed on a child's palms after school; duration: ``within one week.'' Rationale: exact onset and subsequent changes remain unclear. Primary action: time course.\\
\textbf{Example 2:} History: intermittent blurred vision; its specific features are still unclear. Rationale: the specific features of the visual abnormality need clarification. Primary action: characteristics.\\
\textbf{Current decision input $c_t^a$:} State: HPI.\\
History (excerpt): painless, non-itchy rash in spots and patches on the limbs is increasing; topical medication has not relieved it, and the patient asks for medication advice.\\
Doctor: ``How long?'' Patient: ``About twenty days.''\\
Previously queried sub-action: time course / duration.\\
Candidate primary actions: [time course, characteristics, episode pattern, related factors, current consultation, current medication, current non-drug treatment, \texttt{EXIT}].\\
\textbf{Output requirement:} Give a rationale and the selected primary action using only evidence from the current case.}
For the same input we sample $G=7$ completions,
\[
o_i=(e_i,\hat q_i)\sim\pi_\phi(\cdot\mid c_t^q,F_t),\quad i=1,\ldots,G.
\]
An exactly correct parsed label receives one; an incorrect or unparseable label receives zero:
\begin{equation}
R_q(o_i,q_t^*)=\mathbb I\{\operatorname{parse}_q(o_i)=q_t^*\}.
\label{eq:app-choice-reward}
\end{equation}

\textbf{Stage IV: Expansion.}
Using the same prompts, the maker generates one candidate for each of 12,004 primary-action and 13,929 sub-action contexts. These separately sampled pools exclude seed cases but may overlap maker training. We require correct labels, nonempty rationales, and untruncated output. The same checkpoint, prompted as reviewer $V_\phi$, checks grounding, label consistency, and unsupported diagnosis/treatment advice; only \emph{pass} survives, excluding \emph{reject} and \emph{uncertain}.
Thus the accepted set is
\begin{equation}
\begin{split}
\mathcal D_e^q=\{(c_t^q,e,q_t^*):\;&\operatorname{parse}_q(o)=q_t^*,\ e\ne\varnothing,\\
&\operatorname{complete}(o)=1,\\
&V_\phi(c_t^q,q_t^*,e)=\mathrm{pass}\}.
\end{split}
\end{equation}
Here completeness means no truncation. At most one candidate is retained per context, yielding 3,961 primary-action and 4,402 sub-action examples: 8,363 SFT pairs in total.

\textbf{Stage V: Fresh-backbone SFT.}
We initialize a new LoRA adapter on $M_0$ and transfer the maker's knowledge through accepted examples. The input is the current context and the target is rationale plus reference choice. State examples use Eq.~\ref{eq:app-state-loss}; decision examples use
\begin{equation}
\mathcal L_{\mathrm{dec}}(\theta)=-\mathbb E_{(c^q,e,q^*)\sim\mathcal D_e}
\log p_\theta(e,q^*\mid c^q).
\end{equation}
Only rationale and choice tokens contribute to decision loss; inputs and padding are masked, and seed demonstrations are omitted.

\textbf{Stage VI: Final-policy GRPO.}
From the SFT adapter, we refine both decision levels with $G=7$, Eq.~\ref{eq:app-choice-reward}, and the main-text GRPO objective. Inputs supply annotated parent controls and admissible choices, including \texttt{EXIT}, but no seed demonstrations. Selected non-terminal sub-actions are immediately masked; their parent remains available. Current reference labels serve only scoring. Both examples below use the same case and show one completion and reward.
\prismexample{Training example 1: primary action.}{%
\textbf{Input $c_t^a$---history:} Doctor: ``Any discomfort?'' Patient: ``I have been constipated since my kidney biopsy; laxatives have not helped.''\\
Given state: HPI. Candidate actions (excerpt): [time course, characteristics, current medication, \ldots, \texttt{EXIT}].\\
\textbf{Generated output:} $e$: ``Laxatives have not helped, but medication details remain unclear; ask about current medication.'' Selected primary action: current medication.\\
\textbf{Scoring (not input):} Reference $a_t^*$: current medication. Matching label; reward $=1$.}
\prismexample{Training example 2: sub-action.}{%
\textbf{Input $c_t^{a_{\mathrm{sub}}}$---history:} Doctor: ``Any discomfort?'' Patient: ``I have been constipated since my kidney biopsy; laxatives have not helped.''\\
Given state: HPI. Given primary action: current medication. Candidate sub-actions (excerpt): [drug name, \ldots, \texttt{EXIT}].\\
\textbf{Generated output:} $e$: ``The patient mentions a laxative without naming it; first identify the current drug.'' Selected sub-action: drug name.\\
\textbf{Scoring (not input):} Reference $a_{\mathrm{sub},t}^*$: drug name. Matching label; reward $=1$.}

\subsubsection{Supervised Response Realization}
This stage verbalizes the selected information need as a patient-facing question. We train a separate Qwen3-8B LoRA response model on 12,917 training and 642 validation questions. History and annotated state/action controls condition the reference question, $(h_t,s_t^*,a_t^*,a_{\mathrm{sub},t}^*)\mapsto y_t^*$. These controls identify the information to request, and the model formulates the question in context. Training uses annotated parent controls; inference uses the MAP state from belief estimation and actions selected by the policy. Only question tokens and EOS receive cross-entropy supervision; inputs and padding are masked. We use AdamW for three epochs and select minimum validation loss.
\prismexample{Response-training example.}{%
\textbf{Input $h_t$:} Doctor: ``Any discomfort?'' Patient: ``I have been constipated since my kidney biopsy; laxatives have not helped.''\\
\textbf{Given controls:} State $s_t^*$: HPI; primary action $a_t^*$: current medication; sub-action $a_{\mathrm{sub},t}^*$: drug name.\\
\textbf{Target question $y_t^*$:} ``What laxative are you taking now?''}

\subsection{Inference with Different Opening Roles}
Online consultations may begin with a physician using existing records or with a patient describing the chief complaint. PRISM accommodates both openings and then follows the same belief-estimation, decision, and response process.
\par\smallskip
\begingroup
\setlength{\fboxsep}{5pt}
\setlength{\fboxrule}{0.4pt}
\definecolor{prismnote}{RGB}{26,76,100}
\noindent\begin{minipage}{\linewidth}
\colorbox{black}{\parbox{\dimexpr\linewidth-2\fboxsep\relax}{\color{white}\small\bfseries (a) Doctor-Initiated Dialogue}}
\par\nointerlineskip
\fcolorbox{black}{black!3}{\begin{minipage}{\dimexpr\linewidth-2\fboxsep-2\fboxrule\relax}
\normalsize\setlength{\parindent}{0pt}\setlength{\parskip}{1pt}
\colorbox{black!9}{\parbox{\dimexpr\linewidth-2\fboxsep\relax}{\footnotesize
\textbf{Clinical Background.} Stage II membranous nephropathy; prior glucocorticoids and cyclosporine, currently tacrolimus and Chinese medicines; recent diarrhea.}}
\par
{\color{prismnote}\footnotesize\textbf{PRISM:} HPI $\to$ characteristics $\to$ severity.\par}
\textbf{PRISM (Doctor):} How severe is the diarrhea at present?

\textbf{USER (Patient):} I have had severe diarrhea recently. Have tacrolimus side effects worsened? I cannot travel to Beijing. What tests should I have locally? Can I use the treatment plan you mentioned?

{\color{prismnote}\footnotesize\textbf{PRISM:} HPI $\to$ current medication $\to$ drug name.\\
\textbf{Retain concern:} Tests and treatment plan.\par}
\textbf{PRISM (Doctor):} Which medications are you currently taking?

\textbf{USER (Patient):} Tacrolimus costs me over RMB~2,000 a month. I also take antihypertensives and Chinese medicines: over RMB~4,000 monthly, over RMB~50,000 yearly in total.

{\color{prismnote}\footnotesize\textbf{Update:} Add antihypertensives to medication history; retain cost information.\par}
\end{minipage}}
\end{minipage}\par\medskip
\noindent\begin{minipage}{\linewidth}
\colorbox{black}{\parbox{\dimexpr\linewidth-2\fboxsep\relax}{\color{white}\small\bfseries (b) Patient-Initiated Dialogue}}
\par\nointerlineskip
\fcolorbox{black}{black!3}{\begin{minipage}{\dimexpr\linewidth-2\fboxsep-2\fboxrule\relax}
\normalsize\setlength{\parindent}{0pt}\setlength{\parskip}{1pt}
\textbf{USER (Patient):} After two weeks of medication, I still have thick skin on my feet, especially after bathing. Once peeled off, it grows back in days. What is this? Should I continue the medication? What treatment should I have next?

{\color{prismnote}\footnotesize\textbf{PRISM:} HPI $\to$ time course $\to$ duration.\par}
\textbf{PRISM (Doctor):} How long has this been going on?

\textbf{USER (Patient):} I am anxious. It has not improved for over two years; it is getting worse and keeps spreading.

{\color{prismnote}\footnotesize\textbf{PRISM:} HPI $\to$ current medication $\to$ drug name.\\
\textbf{Retain concern:} Diagnosis and treatment.\par}
\textbf{PRISM (Doctor):} Which medications have you used previously?

\textbf{USER (Patient):} Do Lanmeishu cream and Meike cream contain steroids? You previously prescribed Luomeile, vitamin~A, and urea cream for me.

{\color{prismnote}\footnotesize\textbf{Update:} Record medication history; retain the steroid concern for physician review.\par}
\end{minipage}}
\end{minipage}
\endgroup
\par\medskip

\subsection{Hardware and Inference Settings}
Training uses one Linux node with eight H100 80\,GB GPUs, Qwen3-8B. Rollouts use vLLM with tensor parallelism 2, temperature 1.0, top-$p$ 0.95, and top-$k$ 20. Decision and response models use separate LoRA adapters; Table~\ref{tab:appendix-training-settings} gives the main settings.

\par\addvspace{6pt}
\noindent\begin{minipage}{\linewidth}
\captionof{table}{Main training parameters by stage.}
\label{tab:appendix-training-settings}
\centering\small\setlength{\tabcolsep}{1.5pt}
\begin{tabular*}{\linewidth}{@{\extracolsep{\fill}}lcccc@{}}
\toprule
Parameter & Maker & Dec. SFT & Final RL & Resp. SFT \\
\midrule
Epochs & 1 & 2 & 50 & 3 \\
Batch size & 32 & 64 & 128 & 64 \\
Learning rate & $10^{-5}$ & $5\!\times\!10^{-5}$ & $10^{-4}$ & $5\!\times\!10^{-5}$ \\
LoRA rank $r$ & 32 & 32 & 32 & 32 \\
LoRA scale $\alpha$ & 64 & 64 & 64 & 64 \\
Warm-up ratio & 0.03 & 0.03 & 0.03 & 0.03 \\
Rollouts $G$ & 7 & --- & 7 & --- \\
KL coefficient $\beta$ & 0.001 & --- & 0.001 & --- \\
\bottomrule
\end{tabular*}
\end{minipage}\par\smallskip

\section{Implementation of Interaction Metrics}
\label{app:interaction-metrics}
The procedure implements the Section~3.3 metrics by identifying patient initiative and physician uptake, then averaging dialogue-level rates. Turn pairs below are valid pairs after preprocessing and annotation checks.

\textbf{Stage I: Eligible turns.} Remove empty/media-only messages and merge consecutive same-speaker messages. Evaluate patient input $u_t$ only if a prior patient turn exists and a physician immediately precedes it. As in Section~2, $h_t$ includes $u_t$ and all preceding dialogue; $y_t$ is the following physician reply, if any. Patient labels use $h_t$ without future replies, separate prior information, or source labels; uptake also uses $y_t$. Without $y_t$, only patient behavior can be assessed.

\textbf{Stage II: Patient initiative.} We use Qwen3.8-27B to separately identify two behaviors: disclosure $D_t$, a clinical fact not previously stated in the history and outside active physician inquiries; and redirection $A_t$, an unprompted novel clinical concern introducing a distinct health focus beyond the active inquiry. Unanswered earlier questions and relevant open invitations count as elicitation. Responses to these invitations are not unsolicited disclosures; symptom elaboration or responses to physician-led topic transitions are not redirections. Each label is $1$, $0$, or $\bot$ for yes, no, or uncertain. At least one confirmed behavior establishes Any initiative:
\[
I_t=\begin{cases}
1,&D_t=1\ \text{or}\ A_t=1,\\
0,&D_t=A_t=0,\\
\bot,&\text{otherwise}.
\end{cases}
\]
The model returns reasons and evidence; positive labels require an exact span in $u_t$. Uncertain, failed, or invalid annotations are missing, not negative.

\textbf{Stage III: Physician uptake.} For $I_t=1$ with $y_t$, the model uses initiative labels/evidence and the reply to judge whether the physician substantively incorporates the patient-initiated content into a response or follow-up question. The uptake label $U_t$ is $1$ for confirmed uptake, $0$ for confirmed non-uptake, and $\bot$ when uncertain. Generic acknowledgement does not count. Positive uptake requires a continuous span in $y_t$; matching verifies that the quoted evidence exists, while the annotation model judges its meaning.

\textbf{Stage IV: Dialogue-level rates.} Discl., Redir., and Any are positive proportions among their respective determinate labels and require no subsequent physician reply. Cond. asks how often an initiative is taken up; Colla. asks how often initiative and uptake co-occur in evaluable patient--physician pairs. Let $V_d^U$ contain initiative turns with a follow-up and determinate uptake, and $Z_d$ contain confirmed non-initiative turns with a follow-up:
\[
\mathrm{Cond.}_d=\frac{\sum_{t\in V_d^U}U_t}{|V_d^U|},\quad
\mathrm{Colla.}_d=\frac{\sum_{t\in V_d^U}U_t}{|V_d^U|+|Z_d|}.
\]
Both numerators count initiatives taken up by the physician. Cond. considers only initiative turns; Colla. also counts confirmed non-initiative turns in its denominator, assigning them zero. Initiative turns without a physician follow-up or a determinate uptake label enter neither rate.

\textbf{Stage V: Aggregation.} For metric $M$, $V_d^M$ is its valid turn set, $w_{d,t}^M$ the binary outcome, and $\mathcal D_M$ the dialogues with nonempty valid sets:
\[
\widehat r_M=\frac{1}{|\mathcal D_M|}\sum_{d\in\mathcal D_M}
\frac{\sum_{t\in V_d^M}w_{d,t}^M}{|V_d^M|}.
\]
We average dialogue-level rates with equal weight, omitting dialogues with empty valid sets; longer conversations receive no extra weight. Category summaries equally weight datasets. For percentile 95\% confidence intervals, resample whole groups sharing a case/scenario with replacement and recompute the metric in each bootstrap sample. This preserves within-case/scenario dependence.

\section{Limitations}
\label{app:limitations}
As an AI research team, we would ideally involve clinical experts in independent evaluation and the calibration of medical annotations. However, privacy requirements and partner agreements limit the use of raw consultation records to internal research, constraining opportunities for external clinical review. As a compensatory measure, we use a privately deployed model to annotate each sample’s structured consultation labels three times and retain only samples with complete agreement across all three passes, aiming to improve annotation consistency and stability.

\end{document}